\def\year{2018}\relax
\pdfoutput=1
\documentclass[letterpaper]{article} 
\usepackage{aaai18}  
\usepackage{times}  
\usepackage{helvet}  
\usepackage{courier}  
\usepackage{url}  
\usepackage{graphicx}  
\usepackage{amsfonts}
\usepackage{amsmath}
\usepackage{tabularx, booktabs}
\usepackage{algorithm}
\usepackage[noend]{algpseudocode}
\usepackage{verbatim} 
\makeatletter
\def\maketag@@@#1{\hbox{\m@th\normalfont\normalsize#1}}
\makeatother

\usepackage{flexisym} 
\usepackage{bm} 
\usepackage{enumitem} 
\usepackage{color}

\title{Source Traces for Temporal Difference Learning}
\author{Silviu Pitis\\
Georgia Institute of Technology\\
Atlanta, GA, USA 30332\\
spitis@gatech.edu
}
\begin{document}
\maketitle
\begin{abstract}
This paper motivates and develops source traces for temporal difference (TD) learning in the tabular setting. Source traces are like eligibility traces, but model potential histories rather than immediate ones. This allows TD errors to be propagated to potential causal states and leads to faster generalization. Source traces can be thought of as the model-based, backward view of successor representations (SR), and share many of the same benefits. This view, however, suggests several new ideas. First, a TD($\lambda$)-like source learning algorithm is proposed and its convergence is proven. Then, a novel algorithm for learning the source map (or SR matrix) is developed and shown to outperform the previous algorithm. Finally, various approaches to using the source/SR model are explored, and it is shown that source traces can be effectively combined with other model-based methods like Dyna and experience replay.
\end{abstract}

\section{Introduction}

When we, as humans, first learn that some state of the world is valuable (or harmful), we adjust our behavior. For instance, suppose we burn our hand on a hot pan. At least three types of generalization enable useful learning. First, we generalize to substantially similar causal states: we will be more careful handling hot pans on Tuesdays, even if the burn happened on a Monday. Second, we generalize back through time from direct causes to indirect ones. Rather than simply stopping at the last second, we may now instinctively reach for an oven mitt before approaching the pan. Finally, we generalize not only to actual causes, but also to \textit{potential} causes. Our burn may make us less willing to keep our hand close to a campfire while roasting marshmallows.

Reinforcement learning mirrors human learning in many respects, including the three modes of generalization just discussed. The first---generalizing to substantially similar causes---is a direct result of function approximation (e.g., \citeauthor{mnih2013playing} \citeyear{mnih2013playing}). The latter two---generalization to indirect causes and generalization to potential causes---though aided by function approximation, are driven primarily by temporal difference (TD) learning \cite{sutton1988learning}. TD learning uses a value function to bootstrap learning, allowing new information to flow backward, state-by-state, from effects to causes. The speed of one-step TD methods is limited, however, as they require many repeat experiences to generalize.

Several families of methods achieve faster generalization by propagating TD errors several steps per real experience. Eligibility trace methods keep a short-term memory vector that accumulates the (decaying) ``eligibilities'' of recently visited states, and use it to propagate TD errors to past states in proportion to their eligibility \cite{sutton1988learning}. Experience replay (ER) \cite{lin1992self} and Dyna \cite{sutton1990integrated} methods learn from replayed or generated experiences, so that several value function backups occur per real experience. Unlike eligibility traces, ER and Dyna employ a long-term model to speed up learning with respect to known potential causes.

This paper introduces source traces. Source traces are like eligibility traces, but model potential histories rather than immediate ones. This allows for propagation of TD errors to potential causal states and leads to faster learning. Source traces can be thought of as the model-based, backward view of successor representations (SR) \cite{dayan1993improving}. This backward view suggests several new ideas: a provably-convergent TD($\lambda$)-like algorithm for learning and the concept of partial source traces and SR (Section \ref{section_convergence}), a novel algorithm for learning the source map (Section \ref{section_approximate_source_traces}), the triple model learning algorithm (Section \ref{section_approximate_source_traces}), and new perspectives on the source map, which allow, among other things, the use of source traces to enhance experience replay (Section \ref{section_benefits}).

For clarity and brevity, and as a building block to future work, this paper focuses on the tabular MRP-valuation setting. It is likely, however, given existing extensions of SR (e.g., \citeauthor{barreto2016successor} \citeyear{barreto2016successor}), that source traces can be effectively extended to control settings requiring approximation, which is discussed briefly in the conclusion (Section \ref{section_conclusion}).

\section{Ideal Source Traces} \label{section_formalities}

We consider an $n$-state discounted Markov Reward Process (MRP) with infinite time horizon, formally described as the tuple ($S$, $p$, $r$, $\gamma$), where: $S$ is the state space; $p$ is the transition function, which takes two states, $s, s' \in S$, and returns the probability of transitioning to $s'$ from $s$; $r$ is the reward function, which takes a state, $s$, and returns a real-valued reward for that state; and $\gamma \in [0, 1)$ is the discount factor. If the size of the state space, $\vert S \vert$, is finite, then $p$ can be described by the $\vert S \vert \times \vert S \vert$ probability matrix $\textbf{P}$  whose $ij$-th entry is $p(s_i, s_j)$. Similarly, $r$ can be described by the $\vert S \vert$-dimensional vector $\textbf{r}$ whose $i$-th entry is $\mathbb{E}\{r(s_i)\}$. We seek the value of each state, $v(s)$ for $s \in S$, which is equal to the sum of expected discounted future rewards, starting in state $s$. Letting $\textbf{v}$ be the vector whose $i$-th entry is $v(s_i)$:
\begin{equation}
\begin{split}\label{v_eq_qr_derivation}
\textbf{v} & = \textbf{r} + \gamma\textbf{P}\textbf{r} + \gamma^2\textbf{P}^2\textbf{r} \dots \\
& = \left(\textbf{I} + \gamma \textbf{P} + \gamma^2\textbf{P}^2 \dots \right)\textbf{r}\\
& = \left(\textbf{I} - \gamma \textbf{P}\right)^{-1}\textbf{r}\\
\end{split}
\end{equation}

\noindent where the third equality holds because $\gamma < 1$ and $\textbf{P}$ is a transition matrix, so that $\Vert \gamma \textbf{P} \Vert\ < 1$ and the well known matrix identity applies (see, e.g., Theorem A.1 of \citeauthor{sutton1988learning} \citeyear{sutton1988learning}). Alternatively, note that $\textbf{v} = \textbf{r} + \gamma\textbf{P}\textbf{v}$, from which the result follows by arithmetic, given that $\textbf{I} - \gamma \textbf{P}$ is invertible.

In practice, $\textbf{v}$ is often solved for iteratively, rather than by inverting $\textbf{I} - \gamma \textbf{P}$ (see Section 4.1 of \citeauthor{sutton1998reinforcement} \citeyear{sutton1998reinforcement}). For the moment, however, let us assume that we already have the \textit{source map} $\textbf{S} = \left( \textbf{I} - \gamma \textbf{P} \right)^{-1}$. Then, given any $\textbf{r}$, we can solve for $\textbf{v}$ in a single matrix multiplication.

We define the \textit{ideal source trace} for state $s_j$ as the $j$-th column of $\textbf{S}$, $[\textbf{S}]_{\cdot j}$.  
If we have already computed $\textbf{v}$ for some $\textbf{r}$, and $\mathbb{E}\{r(s_j)\}$ changes,
we can use the source trace to precisely propagate the expected reward delta, $\Delta r_j = \mathbb{E}\{r\textprime(s_j)\} - \mathbb{E}\{r(s_j)\}$, and update $\textbf{v}$ in $O(\vert S \vert)$ time:
\begin{equation}
\textbf{v\textprime} = \textbf{v} + \Delta r_j [\textbf{S}]_{\cdot j}.\\
\end{equation}

This ``source backup'' operation is easily extended to \textit{expected} temporal differences. Given some model $\textbf{v}_0$:
\begin{equation}
\begin{split}\label{eq_full_source_backup}
\textbf{v} & = \textbf{S}\textbf{r} + \textbf{v}_0 - \textbf{v}_0\\
& = \textbf{v}_0 + \textbf{S}\left(\textbf{r} - \textbf{S}^{-1} \textbf{v}_0\right)\\
& = \textbf{v}_0 + \textbf{S}\left(\textbf{r} + \gamma \textbf{P} \textbf{v}_0 - \textbf{v}_0\right).\\
\end{split}
\end{equation}

In words: to arrive at the correct value of $\textbf{v}$ from $\textbf{v}_0$, the expected temporal difference upon leaving each state $s_j$, $[\textbf{r} + \gamma \textbf{P} \textbf{v}_0 - \textbf{v}_0]_{j \cdot}$, needs to be added to $\textbf{v}_0$ in proportion to the values of $[\textbf{S}]_{\cdot j}$, the ideal source trace for $s_j$. 

\subsection{Related Work---SR and LSTD}

The source map is well known and goes by several names. When speaking of undiscounted absorbing Markov chains, the source map $(\textbf{I} - \textbf{P})^{-1}$ is known as the \textit{fundamental matrix} \cite{kemeny1960finite}. In the RL setting, it is equivalent to the matrix of \textit{successor representations} \cite{dayan1993improving}. 

\begin{figure*}[ht]
\includegraphics[width=\linewidth]{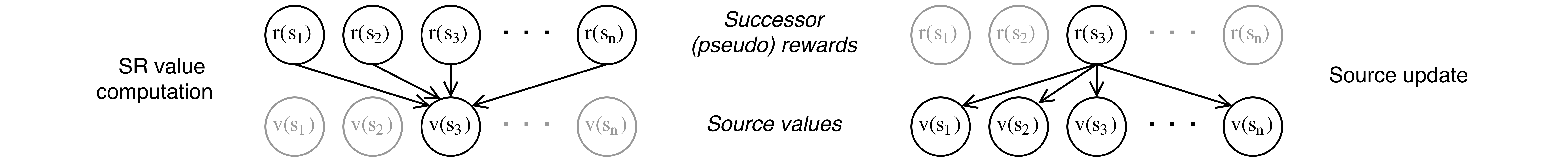}
\caption{SR value computation (left) vs source update (right). When using the SR, one computes $v_0(s_i)$ (as $[\textbf{S}_0]_{i \cdot}\textbf{r}_0$) by looking \textit{forward} to the (pseudo) reward values of its successors, which are learned according to equation \ref{eq_pseudo_rewards} or equation \ref{eq_pseudo_reward_descent}. With source traces, one updates $\textbf{v}_0$ directly by distributing the (pseudo) reward delta at state $s_j$ \textit{backward} to its sources in proportion to $[\textbf{S}_0]_{\cdot j}$ (equation \ref{eq_source_update}). Given a fixed source map, it follows from the derivation of equation \ref{eq_pseudo_rewards} that using the SR with equation \ref{eq_pseudo_rewards} and source learning with equation \ref{eq_source_update} are equivalent. The difference that arises when the source map changes during learning is explored in  Section \ref{section_approximate_source_traces}---Direct Learning vs SR-Reward Decomposition.}
\label{fig_sr_vs_st}
\end{figure*}

Whereas source traces look \textit{backward} to causes, successor representations (SR) look \textit{forward} to effects: the SR of state $s_i$, defined as the $i$-th \textit{row} of $\textbf{S}$, is a $\vert S \vert$-dimensional vector whose $j$-th element can be understood as the cumulative discounted expected number of visits to $s_j$ when starting in $s_i$. While TD learning with source traces updates $\textbf{v}_0$ directly, TD learning with SR learns the model $\textbf{r}_0$ and uses it to compute $v_0(s_i)$ as the dot product $[\textbf{S}_0]_{i \cdot}\textbf{r}_0$ (models are denoted with the subscript 0 throughout). The two perspectives are illustrated in figure \ref{fig_sr_vs_st}. Though different in motivation and interpretation, source traces and successor representations share the matrix $\textbf{S}$ and are useful for similar reasons.

Much like the backward and forward views of TD($\lambda$) offer different insights \cite{sutton1998reinforcement}, so too do the backward view (source traces) and the forward view (SR) here. For instance, the forward view suggests the following useful characterization: we can think of $\textbf{S}_{ij}$ as the solution $v(s_i)$ of a derivative MRP with the same dynamics as the underlying MRP, but with zero reward everywhere \textit{except} state $s_j$, where the reward is 1. This suggests that source traces may be found by TD learning as the solutions to a set of derivative MRPs (one for each state). This algorithm was proposed by \citeauthor{dayan1993improving} \citeyear{dayan1993improving}. By contrast, the backward view prompts an alternative algorithm based on TD($\lambda$) that offers a faster initial learning curve at the price of increased variance. We combine the two algorithms in Section \ref{section_approximate_source_traces} to get the best of both worlds. This shift in perspective produces several other novel ideas, which we explore below. 

Source learning is also related to Least-Squares TD (LSTD) methods \cite{bradtke1996linear}. In the tabular case, LSTD  models $\textbf{S}^{-1}$ (i.e., $\textbf{I} - \gamma\textbf{P}$) and $\textbf{r}$ (or scalar multiples thereof), and computes $\textbf{v}_0$ as $(\textbf{S}_0^{-1})^{-1}\textbf{r}_0$. Incremental LSTD (iLSTD) \cite{geramifard2006incremental} is similar, but learns $\textbf{v}_0$ incrementally, thus avoiding matrix inversion. Finally, recursive LSTD (RLS TD) \cite{bradtke1996linear} computes $\textbf{v}_0$ as does LSTD, but does so by recursive least squares instead of matrix inversion. Of these algorithms, RLS TD---despite being an algorithmic trick---is closest in spirit to source learning; in the tabular case, equation 15 of \citeauthor{bradtke1996linear} and equation \ref{eq_source_update} (below) have comparable structure. However, RLS TD maintains its estimate of $\textbf{v}$ precisely using $O(\vert S \vert^2)$ time per step, whereas the TD Source algorithm presented learns $\textbf{S}_0$ and $\textbf{v}_0$ using $O(\vert S \vert)$ time per step. As iLSTD has the same time complexity as source learning in the tabular case, we compare them empirically in Section \ref{section_approximate_source_traces}.

\section{Convergence} \label{section_convergence}

Equation $\ref{eq_full_source_backup}$ shows that a synchronous source backup of all expected TD errors converges after a single iteration. But what about in the reinforcement learning setting, where rewards and transitions are sampled asynchronously? And what if our model of $\textbf{S}$ is only approximate? 

Unlike the Q-learning backup operator, the asynchronous source backup operator is not generally a contraction in any weighted maximum norm of the value space, even with known $\textbf{S}$. As an example, consider the two state MRP defined by $\textbf{r} = [0, 0]^T$, $\textbf{P} = [[0.5, 0.5], [0.5, 0.5]]$, and $\gamma = 0.5$, so that $\textbf{S} = [[1.5, 0.5], [0.5, 1.5]]$. The optimal $\textbf{v}$ is clearly $[0, 0]^T$, but starting at $\textbf{v}_0 = [a, -a]^T$, the asynchronous backup from the first state (state $i$) results in a TD error of $-a$, so that $\textbf{v}_1 = \textbf{v}_0 + [\textbf{S}]_{\cdot i}\left([r]_i + \gamma[\textbf{P}]_{\cdot i}^T\textbf{v}_0 - [\textbf{v}_0]_i\right) = [a, -a]^T - a[1.5, 0.5]^T = [-0.5a, -1.5a]^T$. It is easy to see that the value of state $j$ has been pushed away from its optimal value by $0.5a$: $\vert [\textbf{v}_1 - \textbf{v}]_j \vert\ > \vert [\textbf{v}_0 - \textbf{v}]_j \vert$. By symmetry, the same expansion in $v(s_i)$ occurs for the asynchronous backup from state $j$. Since $a$ is arbitrary, the conclusion follows.

Nevertheless, by working in reward space, instead of value space, proof of convergence is reduced to a straightforward application of the general convergence theorem of \citeauthor{jaakkola1994convergence} \citeyear{jaakkola1994convergence}. Notably, convergence is guaranteed even with sufficiently approximated $\textbf{S}$.

\noindent \textbf{Theorem.} The sampled, asynchronous source learning algorithm given by:
\begin{small}
\begin{equation} \label{eq_source_update}
\scalebox{0.9}[1.]{$\textbf{v}_{n+1} = \textbf{v}_{n} + \alpha_n[\textbf{S}_0]_{\cdot \langle n \rangle}\bigg(r(s_{\langle n \rangle}) + \gamma v_n(s_{\langle n + 1 \rangle}) - v_n(s_{\langle n \rangle})\bigg)$}
\end{equation}
\end{small}%
\noindent where subscript $\langle n \rangle$ denotes the index corresponding to the state at time $n$ and $\textbf{S}_0$ is an approximation of $\textbf{S}$, converges to the true $\textbf{v}$ with probability 1 if:
\begin{enumerate}[label=(\alph*),leftmargin=2\parindent]
	\item the state space is finite,
	\item $\forall$ state indices $x$, $\sum\limits_{\langle n \rangle = x} \alpha_n = \infty$ and $\sum\limits_{\langle n \rangle = x} \alpha_n^2 < \infty$,
	\item $\text{Var}\{r(s_i)\}$ is bounded for all $i$, and
	\item $\Vert (\textbf{I} - \textbf{S}^{-1}\textbf{S}_0) \Vert\ \leq \gamma$. 
\end{enumerate}

\noindent \textbf{Proof.} Let $\textbf{d}_n$ and $\bm{\alpha}_n$ be $\vert S \vert$-dimensional vectors, where the $i$-th slot of $\textbf{d}_n$ corresponds to a sampled TD error starting in state $s_i$, $r(s_i) + \gamma v_n(s_{next(i)}) - v_n(s_i)$ (all samples but the one corresponding to $s_{\langle n \rangle}$ are hypothetical), and $\bm{\alpha}_n$ is zero everywhere except in the position corresponding to $s_{\langle n \rangle}$, where it is $\alpha_n$. Then, we can rewrite equation \ref{eq_source_update} as:
\begin{equation} \label{proof_value_eq}
\textbf{v}_{n+1} = \textbf{v}_{n} + \textbf{S}_0(\bm{\alpha}_n \odot \textbf{d}_n)
\end{equation}

\noindent where $\odot$ is the element-wise product. 

Since $\Vert (\textbf{I} - \textbf{S}^{-1}\textbf{S}_0) \Vert\ \leq \gamma < 1$, the matrix identity used in equation \ref{v_eq_qr_derivation} above implies that $\textbf{S}^{-1}\textbf{S}_0$ is invertible, so that $\textbf{S}_0$ is invertible. Letting $\textbf{r}_n = \textbf{S}_0^{-1}\textbf{v}_n$ and left-multiplying both sides of (\ref{proof_value_eq}) by $\textbf{S}_0^{-1}$ we obtain the following equivalent update in (pseudo) reward space:
\begin{equation} \label{eq_pseudo_rewards}
\textbf{r}_{n+1} = \textbf{r}_{n} + \bm{\alpha}_n \odot \textbf{d}_n.
\end{equation}
Subtracting $\textbf{S}_0^{-1}\textbf{S}\textbf{r}$ from both sides and rearranging, we obtain the iterative process $\{\bm{\Delta}_n\}$:
\begin{equation} \label{proof_jaakk_format}
\bm{\Delta}_{n+1} = (1 - \bm{\alpha}_n) \odot \bm{\Delta}_n + \bm{\alpha}_n \odot \textbf{F}_n
\end{equation}
\noindent where $\bm{\Delta}_n = \textbf{r}_n - \textbf{S}_0^{-1}\textbf{S}\textbf{r}$ and $\textbf{F}_n = \textbf{d}_n + \bm{\Delta}_n$. To this process, we apply Theorem 1 of \citeauthor{jaakkola1994convergence} \citeyear{jaakkola1994convergence}. Conditions (1) and (2) of \citeauthor{jaakkola1994convergence} are satisfied by the assumptions above. Condition (3) is satisfied because:
\begin{equation}
\begin{split}
\Vert \mathbb{E}\{\textbf{F}_n\} \Vert\ & = \Vert \textbf{r} - \textbf{S}^{-1}\textbf{S}_0\textbf{r}_n + \bm{\Delta}_n \Vert \\
& = \Vert (\textbf{I} - \textbf{S}^{-1}\textbf{S}_0) \bm{\Delta}_n \Vert\\
& \leq  \gamma \Vert \bm{\Delta}_n \Vert.
\end{split}
\end{equation} 

\noindent Condition (4) is satisfied because $[\textbf{F}_n]_k$ depends at most linearly on $[\textbf{r}_{n}]_k$ and $\text{Var}\{r(s_k)\}$ is bounded by assumption. Thus, Theorem 1 of \citeauthor{jaakkola1994convergence} applies, $\{\bm{\Delta}_n\}$ converges to $\textbf{0}$, and so too does $\{\textbf{v}_n - \textbf{v}\}$. 

\subsection{Partial Source Traces and Speed of Convergence}

\begin{figure*}[ht]
\includegraphics[width=\textwidth]{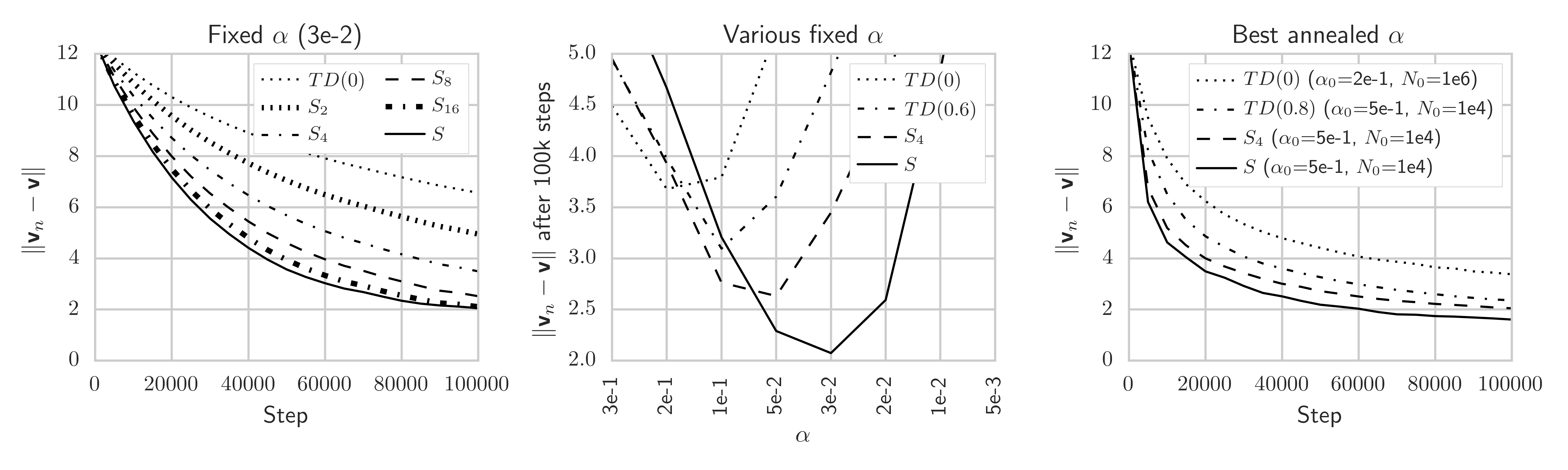}
\caption{Convergence in 3D Gridworld using known partial and full source traces for various learning rate schedules.}
\label{fig_convergence}
\end{figure*}

The theorem admits a variety of approximate models, so long as $\Vert (\textbf{I} - \textbf{S}^{-1}\textbf{S}_0) \Vert\ \leq \gamma$. One such family is that of \textit{$n$-step} source traces (cf. the $n$-step methods of \citeauthor{sutton1998reinforcement} \citeyear{sutton1998reinforcement}), derived from the source maps $\textbf{S}_n = \sum_{k=0}^{n-1}(\gamma\textbf{P})^k$ for $n \geq 1$. Using $\textbf{S}_1 = \textbf{I}$ corresponds to TD(0), so that $n > 1$ interpolates between TD(0) and full source learning. Another way to interpolate between TD(0) and full source learning is to introduce a parameter $\lambda \in [0, 1]$ (cf. $\lambda $ in TD($\lambda$)) and use $\lambda$ source traces, derived from $\textbf{S}^{\lambda} = \sum_{k=0}^{\infty} (\gamma\lambda\textbf{P})^k$. 

We may combine $\textbf{S}_n$ and $\textbf{S}^{\lambda}$ to form:
\begin{equation}
\textbf{S}^{\lambda}_n = \sum_{k = 0}^{n-1} (\gamma\lambda\textbf{P})^{k}.
\end{equation}

\noindent This gives us the more general family of \textit{partial} source traces, for which convergence holds at all $\lambda$ and $n$. The case $n = 1$ is easy to verify. For $n > 1$, we have:
\begin{small}
\begin{equation}
\scalebox{0.9}[1.]{$\begin{aligned}
\Vert \textbf{I} - \textbf{S}^{-1}\textbf{S}^{\lambda}_n \Vert & =
\Vert \textbf{I} - (\textbf{I} - \gamma\textbf{P})\sum_{k = 1}^{n}(\gamma\lambda\textbf{P})^{k-1} \Vert\\
& \hspace*{-30pt} = \Vert (1-\lambda)\gamma\textbf{P}\sum_{k=0}^{n-2}(\lambda\gamma\textbf{P})^{k} + \lambda^{n-1}\gamma^n\textbf{P}^n\Vert\\
& \hspace*{-30pt} \leq \gamma \left[(1 - \lambda)\sum_{k=0}^{n-2}(\lambda\gamma)^k + (\lambda\gamma)^{n-1} \right]\\
& \hspace*{-30pt} = \gamma \left[(1 - \lambda\gamma)\sum_{k=0}^{n-2}(\lambda\gamma)^k + (\lambda\gamma - \lambda)\sum_{k=0}^{n-2}(\lambda\gamma)^k +  (\lambda\gamma)^{n-1} \right]\hspace{-20pt}\\
& \hspace*{-30pt} = \gamma \left[ 1 - \lambda(1 - \gamma)\sum_{k=0}^{n-2}(\lambda\gamma)^k \right] < \gamma\\
\end{aligned}$}
\end{equation}
\end{small}

\noindent where the first inequality holds by a combination of the triangle inequality, the identity $\Vert a\textbf{A} \Vert\ \leq \vert a \vert \Vert \textbf{A} \Vert$, and the fact that all terms are positive. The bracketed quantity on the final line is in $[0, 1]$ and strictly decreasing in both $n$ and $\lambda$, which suggests that the speed of convergence increases as we move from TD(0) (either $n = 1$ or $\lambda = 0$) to full source traces ($n = \infty$ and $\lambda = 1$). 

Experiments in two environments confirm this, with an important caveat. The first environment is a Random MRP with 100 states, 5 random transitions per state with transition probabilities sampled from $\mathcal{U}(0, 1)$ and then normalized (transitions are re-sampled if the resulting $\textbf{P}$ is not invertible), a random reward for each state  sampled from $\mathcal{N}(0, 1)$, and $\gamma = 0.9$. The second is a 1000-state 3D Gridworld with wraparound: states are arranged in a 10$\times$10$\times$10 3D grid, with state transitions between adjacent states (6 per state, including edge states, which wraparound). Transition probabilities are sampled from $\mathcal{U}(0, 1)$ and then normalized. 50 random states are assigned a random reward sampled from  $\mathcal{N}(0, 1)$ and $\gamma$ is set to 0.95.

Unless otherwise noted, similar results held in both environments, here and throughout. All experiments, unless otherwise noted, reflect average results on 30 Random MRP or 3D Gridworld environments. The same set of randomly generated environments was used when comparing algorithms. 

Figure \ref{fig_convergence} (left), which reflects the 3D Gridworld, plots learning curves after 100,000 steps for the source learning algorithm given by equation \ref{eq_source_update} for $\textbf{S}_1$ (TD(0)) through $\textbf{S}_4$ and $\textbf{S}$. A similar pattern appears for $\textbf{S}^{\lambda}$ with increasing $\lambda$. In each case, $\textbf{v}_0$ was initialized to $\textbf{0}$, and $\Vert \textbf{v}_n - \textbf{v} \Vert$ was averaged across the MRPs ($\textbf{v}$ was computed by matrix inversion). 

The curves in figure \ref{fig_convergence} (left) are not representative of all learning rates. Figure \ref{fig_convergence} (center) shows the final error achieved by  TD(0), TD($\lambda$) at the best $\lambda$ (tested in 0.1 increments), $\textbf{S}_4$ and $\textbf{S}$ at various fixed $\alpha$. Although the advantage of full source traces is clear, both TD($\lambda$) and partial source learning do better at higher $\alpha$s. In the Random MRP environment (not shown), partial traces provide the best results after only 50,000 steps and experiments suggest that TD(0) eventually overtakes full source learning at \textit{all} fixed $\alpha$. 

This highlights a weakness of using source traces: they amplify the per-step transition variance by propagating it multiple steps. We propose two strategies for managing this. The first is to use a forward model, which we explore in the ``triple model learning'' algorithm of Section \ref{section_approximate_source_traces}. The second is to anneal the learning rate. For consistent comparison with iLSTD in Section \ref{section_approximate_source_traces}, we tested the following set of annealing schedules, adapted from \citeauthor{geramifard2007ilstd} \citeyear{geramifard2007ilstd}:
\begin{equation}
\alpha_n = \alpha_0(N_0 + 1)/(N_0 + n^{1.1})
\end{equation}

\noindent for $\alpha \in$ \{5e-1, 2e-1, 1e-1, 5e-2, 2e-2, 1e-2, 5e-3\} and $N_0 \in$ \{0, 1e2, 1e4, 1e6\}. Figure \ref{fig_convergence} (right) plots the best results for TD(0), TD($\lambda$) at the best $\lambda$, $\textbf{S}_4$ and $\textbf{S}$ when annealing the learning rate, which improved both the convergence speed and final results for all methods.

\subsection{Relating the Convergence Results to SR}

It is tempting to interpret the convergence theorem in terms of SR by noting that $\textbf{S}_0$ maps to the SR of a discrete state space ($\textbf{S}_0 \Leftrightarrow \bm{\Phi}$), so that the pseudo-reward space of equation \ref{eq_pseudo_rewards} maps to the weight space when using the SR with linear value function approximation ($\textbf{r}_n \Leftrightarrow \bm{\theta}_n$). NB, however, that equation \ref{eq_pseudo_rewards} updates only a single element of $\textbf{r}_n$, whereas semi-gradient TD(0) over $\bm{\theta}$, which was the rule used in \citeauthor{dayan1993improving} \citeyear{dayan1993improving} and which we term \textit{pseudo-reward descent}, updates (potentially) all elements of $\bm{\theta}_n$ with the rule:
\begin{equation} \label{eq_pseudo_reward_descent}
\bm{\theta}_{n+1} = \bm{\theta}_{n} + \alpha_n \delta_n \bm{\phi}(s_{\langle n\rangle})
\end{equation}
\noindent where $\delta_n = [\textbf{d}_n]_{\langle n\rangle}$ is the temporal difference at step $n$ and $\bm{\phi}(s_{\langle n\rangle})$ is the SR of $s_{\langle n \rangle}$. Although the convergence of (\ref{eq_pseudo_reward_descent}) does not follow from our theorem, it is guaranteed, under slightly different conditions, by the theorem of \citeauthor{tsitsiklis1997analysis} \citeyear{tsitsiklis1997analysis}. A notable difference is that the theorem of \citeauthor{tsitsiklis1997analysis} requires updates to be made according to the invariant distribution of the MRP. We compare the performance of these two update rules in Section \ref{section_approximate_source_traces}.

Setting aside the above differences, we note that $\textbf{S}^{\lambda}_n$ may be used to define partial successor representations (i.e., SR with $n$-step or decaying lookahead) and conjecture that the speed of pseudo-reward descent will increase with $n$ and $\lambda$, as it did for source learning. This idea can be related to the work of \citeauthor{gehring2015approximatesr} \citeyear{gehring2015approximatesr}, which shows that using low-rank approximations of $\textbf{S}$ can obtain good results. Note that Gehring's low-rank approximations are quite different from the invertible approximations presented here and that the two methods for approximation might be combined. 

\section{Learning and Using the Source Map} \label{section_approximate_source_traces}

Source traces are not useful without a practical method for computing them. In this section we draw inspiration from TD($\lambda$) to show that the source model can be learned. While our convergence theorem does not formally apply to this scenario (since $\textbf{S}_0$ is changing), we conjecture that convergence will occur if the changes to $\textbf{S}_0$ are sufficiently ``small''.

\subsection{The Link Between Source and Eligibility Traces}

The TD($\lambda$) algorithm \cite{sutton1988learning}, defined by the update:
\begin{small}
\begin{equation} \label{eq_tdlam_update}
\scalebox{0.95}[1.]{$\textbf{v}_{n+1} = \textbf{v}_{n} + \alpha_n \textbf{e}_n \bigg(r(s_{\langle n \rangle}) + \gamma v_n(s_{\langle n + 1 \rangle}) - v_n(s_{\langle n \rangle})\bigg)$}
\end{equation}
\end{small}%
\noindent propagates the TD error of each experience according to the \textit{eligibility trace} at time $n$, $\textbf{e}_n$, defined recursively as:
\begin{equation} \label{elig_recursive_def}
\textbf{e}_n = [\textbf{I}]_{\cdot \langle n \rangle} + \gamma \lambda \textbf{e}_{n-1}
\end{equation}
\noindent where subscript $\langle n \rangle$ denotes the index corresponding to $s_{\langle n \rangle}$. 

It notable that, but for replacing the current eligibility trace with the source trace for the current state, the asynchronous source learning update (equation \ref{eq_source_update}) is identical to the TD($\lambda$) update (equation \ref{eq_tdlam_update}). This suggests a link between source traces and eligibility traces.

Indeed, it is \textit{almost} correct to suppose that the average eligibility trace given that the current state is $s_j$ is the source trace for $s_j$. But recall from equation $\ref{v_eq_qr_derivation}$ that $\textbf{S} = \textbf{I} + \gamma \textbf{P} + \gamma^2\textbf{P}^2 \dots = \textbf{I} + \gamma\textbf{P}\textbf{S}$, so that the source trace for $s_j$ is:
\begin{equation} \label{source_trace_recursive_def}
[\textbf{S}]_{\cdot j} = [\textbf{I}]_{\cdot j} + \gamma \sum_i p(s_i, s_j) [\textbf{S}]_{\cdot i}
\end{equation}
\noindent In contrast, averaging the eligibility traces for state $s_j$ would reverse the roles of $s_j$ and $s_i$; letting $\overline{\textbf{e}}_j$ represent the average eligibility trace in state $s_j$, it is easy to see that:
\begin{equation}
\begin{split}
\overline{\textbf{e}}_j =\ & \mathbb{E}\{[\textbf{I}]_{\cdot j} + \gamma \lambda \textbf{e}_{i}\}\\
=\ & [\textbf{I}]_{\cdot j} + \gamma \lambda \sum_i P(s_i | s_j) \overline{\textbf{e}}_i.\\
\end{split}
\end{equation}
\noindent Importance sampling at each step of the accumulation corrects for this and turns what would otherwise be the average eligibility trace into an estimator of the source trace. It is reflected in the algorithm below (line 12). 

\subsection{The Tabular TD Source Algorithm}

The TD Source algorithm is similar to TD($\lambda$), except that it uses the observed history to update the model $\textbf{S}_0$ rather than to directly distribute TD errors. This adds extra computation per step, and requires storing an $\vert S \vert \times \vert S \vert$ matrix in memory. Line 12 of the algorithm corresponds to equation $\ref{source_trace_recursive_def}$ above. $\beta$ is the learning rate used in the stochastic approximation of $\textbf{S}$ on line 13, and may be fixed or annealed according to some schedule. Lines 14 and 15 are commented out and are not part of the TD Source algorithm (see next subsection). 

\begin{algorithm} \label{algorithm}
\caption{Tabular TD learning with source traces}
\begin{algorithmic}[1]
\Procedure{TD Source[-sr]}{\textit{episodes}, $\gamma$, $\lambda$, $\alpha$, $\beta$}
\State \textbf{v} $\gets$ \textbf{0}
\State $\textbf{S}$ $\gets$ $\textbf{I}$ ($\vert S \vert \times \vert S \vert$ identity matrix)
\State \textbf{c} $\gets$ \textbf{0} (counts vector)
\For{episode in 1 .. \textit{n}}
\State $j \gets$ index of initial state
\State $c(s_j) \gets c(s_j) + 1$
\State $[\textbf{S}]_{\cdot j} \gets (1 - \beta) [\textbf{S}]_{\cdot j} + \beta [\textbf{I}]_{\cdot j}$ 
\For{pair $(s_i, s_{j})$ and reward $r$ in episode}
\State $\textbf{v} \gets \textbf{v} + \alpha [\textbf{S}]_{\cdot i} (r + \gamma v_j - v_i)$
\State $c(s_j) \gets c(s_j) + 1$
\State $\textbf{e} \gets [\textbf{I}]_{\cdot j} + \gamma \lambda \frac{c(s_j)}{c(s_i)} [\textbf{S}]_{\cdot i}$
\State $[\textbf{S}]_{\cdot j} \gets (1 - \beta) [\textbf{S}]_{\cdot j} + \beta\textbf{e}$
\item[] \hspace{\algorithmicindent}\hspace{\algorithmicindent}\hspace{\algorithmicindent}\texttt{/*}
\State $\textbf{e} \gets [\textbf{I}]_{i \cdot} + \gamma \lambda [\textbf{S}]_{j \cdot}$ \Comment{TD SR}
\State $[\textbf{S}]_{i \cdot} \gets (1 - \beta) [\textbf{S}]_{i \cdot} + \beta\textbf{e}$\Comment{TD SR}
\item[] \hspace{\algorithmicindent}\hspace{\algorithmicindent}\hspace{\algorithmicindent}\texttt{*/}
\EndFor
\EndFor\\
\Return \textbf{v}, \textbf{S}, \textbf{c}
\EndProcedure
\end{algorithmic}
\end{algorithm}

The $\lambda$ parameter is included should partial source traces be more desirable than full ones (e.g., because they propagate less variance, or because they are easier to approximate). One might also consider annealing $\lambda$ to 1 in order to minimize the effect of inaccuracies during learning whilst still obtaining a full source model, which we investigate briefly in the next subsection.

\subsection{TD Source vs Dayan's Algorithm (``TD SR'')}

As noted in Section \ref{section_formalities}, the source map is equivalent to the matrix of successor representations, for which there is an existing learning algorithm due to  \citeauthor{dayan1993improving} \citeyear{dayan1993improving} (when combined with equation \ref{eq_source_update}, ``TD SR''). Whereas TD Source learns the source map using the column-wise recurrence of equation \ref{source_trace_recursive_def}, TD SR uses the row-wise recurrence:
\begin{equation} \label{source_trace_rowwise_recursive_def}
[\textbf{S}]_{i \cdot} = [\textbf{I}]_{i \cdot} + \gamma \sum_j p(s_i, s_j) [\textbf{S}]_{j \cdot}.
\end{equation}

TD SR uses lines 14 and 15 of the algorithm in place of lines 12 and 13. And since it does not require importance sampling, lines 4, 6, 7, 8 and 11 can be removed.

As importance sampling often increases variance and slows learning, we might expect TD SR to outperform TD Source. However, the column-wise algorithm has one major advantage: it immediately puts to use the most recent information. For this reason, we should expect TD Source to outperform TD SR algorithm at the start of learning. 

\begin{figure}[ht]
\includegraphics[width=\linewidth]{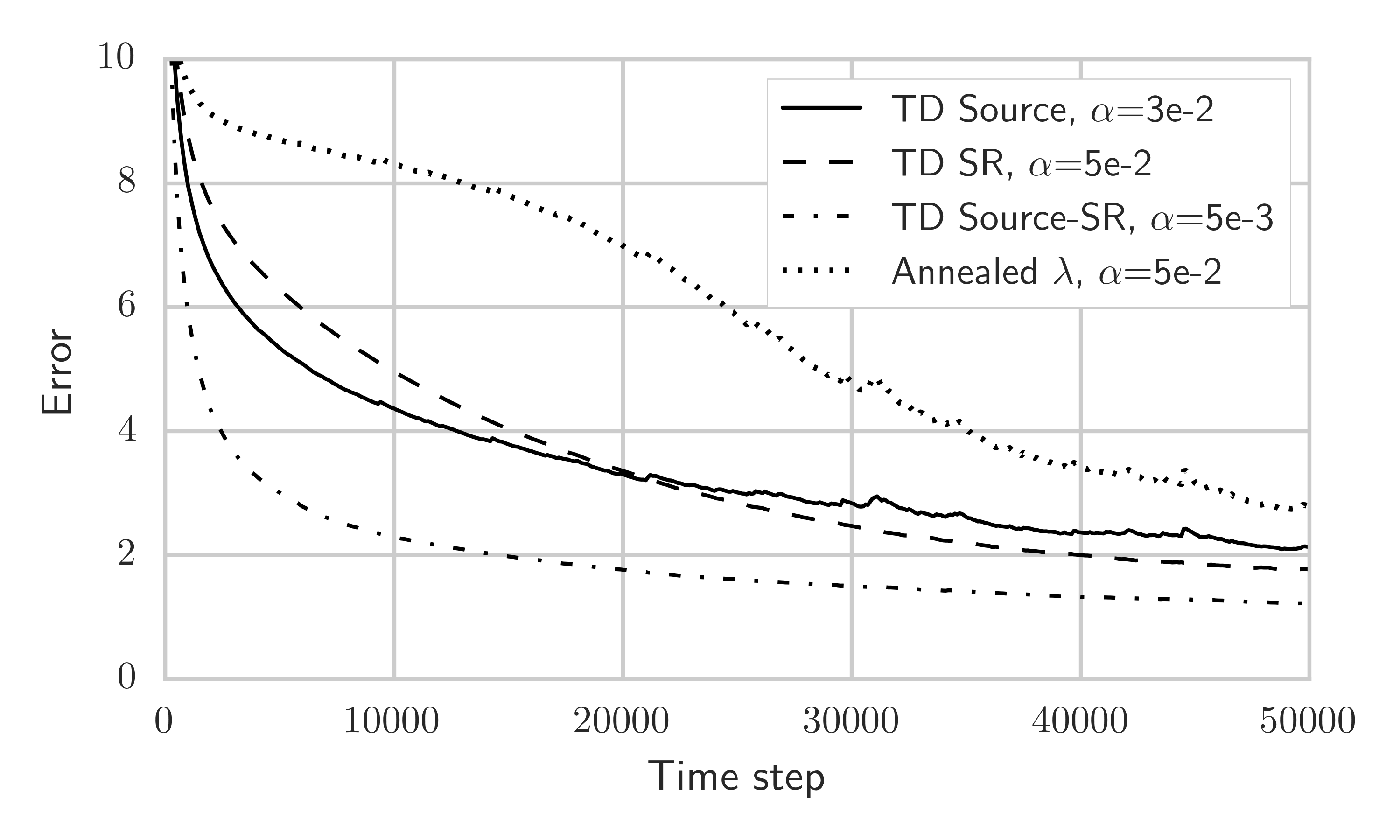}
\caption{TD Source vs TD SR on Random MRPs}
\label{fig_td_source_vs_dayan}
\end{figure}

Experiment confirms this hypothesis. Figure \ref{fig_td_source_vs_dayan} tracks the quality of the approximation over the first 50,000 steps of the Random MRP environment. Four different algorithms were run at various minimum learning rates. The learning curves at the best learning rate (in terms of final error) are shown. For all algorithms, $\alpha$ was annealed to the minimum, on a per-state basis according to the harmonic series, as this performed better in all cases than using fixed $\alpha$. Error was computed as $\sqrt{\sum([\textbf{S}_0 - \textbf{S}]_{ij})^2}$.  

The results suggest that we might improve upon TD Source by either (1) finding a better method for annealing $\alpha$ in response to large importance sampling ratios, or (2) switching to TD SR after an initial training period (neither shown). Alternatively, we might combine TD Source and TD SR and update both $[\textbf{S}]_{\cdot j}$ and $[\textbf{S}]_{i \cdot}$ at each step by using lines 12-15 simultaneously (``TD Source-SR''). This outperforms either algorithm individually by a sizable margin. To check that this is not simply coincidence resulting from a higher effective learning rate, we tested several other learning rate schedules for both TD Source and TD SR individually---none were as effective as TD Source-SR. 

Finally, we tested the proposition in the previous subsection that annealing $\lambda$ to 1 might allow for better approximation of $\textbf{S}$. Starting at $\lambda = 0.5$, $\lambda$ was annealed linearly to 1 over the first 25,000 steps. This approach failed. 

In the 3D Gridworld environment (not shown), similar relationships held after 200,000 steps, although the impact of importance sampling was not as great (due to the two-way local structure) and TD SR did not overtake TD Source. TD Source-SR once again performed best.

\subsection{Direct Learning vs SR-Reward Decomposition}

In Section \ref{section_convergence} we identified three update rules: source learning (equation \ref{eq_source_update}), the equivalent update in pseudo-reward space (equation \ref{eq_pseudo_rewards}), and pseudo-reward descent (equation \ref{eq_pseudo_reward_descent}). When $\textbf{S}_0$ changes during learning, equations \ref{eq_source_update} and \ref{eq_pseudo_rewards} lead to slightly different results, since the former accumulates $\textbf{v}_0$ and is less sensitive to changes in $\textbf{S}_0$. The three rules thus define distinct algorithms, which we term \textit{direct methods}. We call the algorithm corresponding to equation \ref{eq_pseudo_rewards} \textit{White's algorithm}, as it was first proposed by \citeauthor{white1996temporal} \citeyear{white1996temporal} (\S{6.2.1}, with $\lambda = 0$). 

There exists a fourth algorithm. Rather than learn $\textbf{v}$ directly, one may decompose learning into two independent problems: learning $\textbf{S}_0$ and learning $\textbf{r}_0$. Then $\textbf{v} \approx \textbf{S}_0 \cdot \textbf{r}_0$. This LSTD-like approach is taken by \citeauthor{kulkarni2016deep} \citeyear{kulkarni2016deep}. We refer to it as \textit{SR-reward decomposition}. 

Though \citeauthor{white1996temporal} \citeyear{white1996temporal} proposed both White's algorithm and SR-reward decomposition, a comparison of direct learning and decomposition has been lacking. Which is better? 

On one hand, \citeauthor{white1996temporal}  noted that decomposition ``will be out of sync until learning is complete'' and ``as a result, the estimate [of $\textbf{v}$] may be quite poor.'' One may formalize this statement by letting $\bm{\mathcal{E}} = \textbf{S}_0 - \textbf{S}$, $\bm{\epsilon} = \textbf{r}_0 - \textbf{r}$, and putting:
\begin{equation}
\begin{split} \label{eq_decomposition_error}
\textbf{v}_0 - \textbf{v} & = [\textbf{S} + \bm{\mathcal{E}}][\textbf{r} + \bm{\epsilon}] - \textbf{S}\textbf{r}\\
& = \bm{\mathcal{E}}\textbf{r} + \textbf{S}\bm{\epsilon} + \bm{\mathcal{E}}\bm{\epsilon}.\\
\end{split}
\end{equation}
\noindent Then note that the errors compound in the final term. 

On the other hand, $\textbf{r}$ may be easy to learn, so that the latter two terms go to zero and only the error introduced by $\bm{\mathcal{E}}\textbf{r}$ is left. But if $\bm{\mathcal{E}}$ does not go to zero, as in the case of learning a partial source map, then the decomposition approach will not converge to the correct $\textbf{v}$.

Since direct methods are forgiving of inaccurate $\textbf{S}_0$, we may hypothesize that their performance will improve relative to decomposition as the source map becomes harder to learn. We tested this in the 3D Gridworld environment, and also took the opportunity to compare the three direct methods (no parallel experiments were done on Random MRPs).

To modulate the difficulty of learning $\textbf{S}$, we varied the $\gamma$ parameter. This increases the following bound on $\Vert \textbf{S} \Vert$:
\begin{equation}
\Vert \textbf{S} \Vert = \Vert \textbf{I} + \gamma\textbf{P} + \gamma^2\textbf{P}^2 \dots \Vert \leq 1/(1-\gamma).
\end{equation}
\noindent Intuitively, larger $\gamma$ means that rewards are propagating back further, which means larger $\textbf{S}$ and more room for error $\bm{\mathcal{E}}$. 

\begin{figure}[ht]
\includegraphics[width=\linewidth]{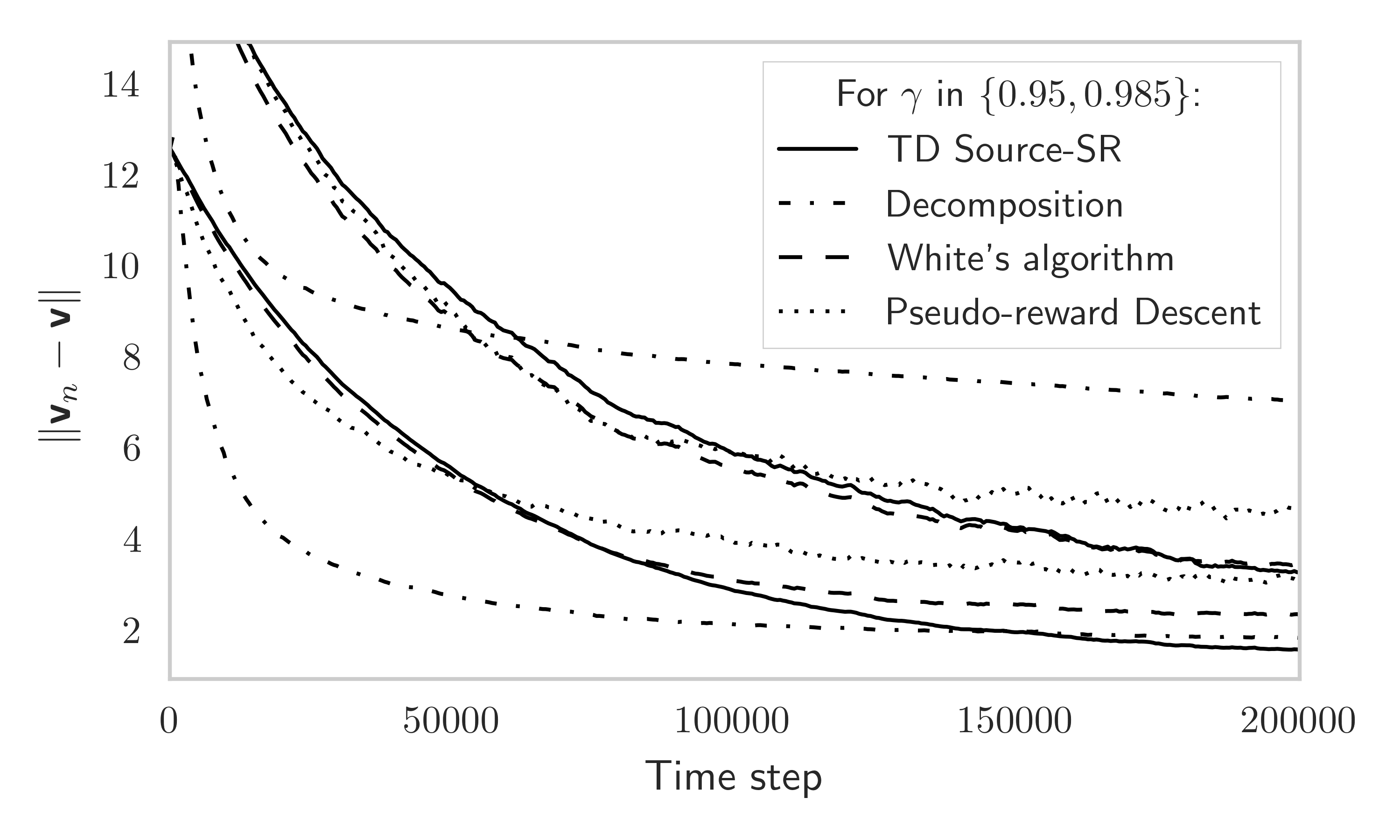}
\caption{Source learning vs SR-reward decomposition vs pseudo-reward descent (higher lines corresp. to higher $\gamma$)}
\label{fig_td_source_vs_decomposition}
\end{figure}

Figure \ref{fig_td_source_vs_decomposition} plots $\Vert \textbf{v}_n - \textbf{v} \Vert$ over the first 200,000 steps in 3D Gridworld at two $\gamma$ values for the four algorithms discussed, each at their best tested fixed $\alpha$, with higher error lines corresponding to higher $\gamma$. For each algorithm, the model of $\textbf{S}$ was learned in the style of TD Source-SR, as described in the previous subsection, so that only the method for computing and updating $\textbf{v}$ was varied.

The results show that decomposition results in faster initial learning in all cases and maintains its advantage for longer the smaller $\gamma$ is. As hypothesized, however, it is overtaken by direct methods as learning progresses. 

\subsection{Triple Model Learning}

The results of the previous subsection suggest the following hybrid strategy: use decomposition at the start of learning and then switch to a direct method. While not unreasonable, it is unclear how to design a switching criterion for such an approach. Instead, we propose the following alternative hybrid, which might be viewed as an instance of Dyna: in addition to learning the backward source model and the reward model, learn a forward transition model; then propagate expected temporal differences rather than sampled ones. This entails the following update:
\begin{small}
\begin{equation*}
\textbf{v}_{n+1} = \textbf{v}_n + [\textbf{S}_0]_{\cdot \langle n \rangle}\left([\textbf{r}_0]_{\langle n \rangle} + \gamma([\textbf{P}_0]_{\langle n \rangle \cdot})^T \textbf{v}_n - v_n(s_{\langle n \rangle})\right)
\end{equation*}
\end{small}%
\noindent where $\textbf{P}_0$ models the transition matrix $\textbf{P}$. 

This update uses three separate models and takes O($\vert S \vert$) time per step. Learning $\textbf{P}_0$ is straightforward in the tabular setting and, like learning $\textbf{S}_0$ and $\textbf{r}_0$, takes O($\vert S \vert$) time per step. We refer to the resulting algorithm as \textit{triple model learning}. Note that since experiences are used solely for learning the models, and learning $\textbf{v}_0$ is strictly model-based, updates may be strategically ordered a la prioritized sweeping \cite{moore1993prioritized}. Note further that since $\textbf{P}_0$ models $p(s_i, s_j)$, it may be used in place of experience when learning $\textbf{S}_0$, which should reduce variance and eliminate the need for importance sampling. 

\subsection{TD Source-SR vs Triple Model Learning vs iLSTD}

In this section we compare TD Source-SR, triple model learning and iLSTD in the 3D Gridworld environment, using the same setup and set of annealing schedules (plus fixed $\alpha$s) used in figure \ref{fig_convergence} (right). iLSTD was chosen for comparison because, like TD Source-SR and triple model learning, it is a model-based method that runs in O($\vert S \vert$) time per step in the tabular setting. We ran both the random and greedy variants of iLSTD, with $m=1$ dimensions updated per iteration, as in \citeauthor{geramifard2007ilstd} \citeyear{geramifard2007ilstd}.\footnote{We note that iLSTD, primarily its greedy variant, ``blew up'' for several annealing schedules. This was the result of accumulating a large $\textbf{A}$. In these cases, we used our knowledge of the true $\textbf{v}$ to terminate learning early.}

\begin{figure}[ht]
\includegraphics[width=\linewidth]{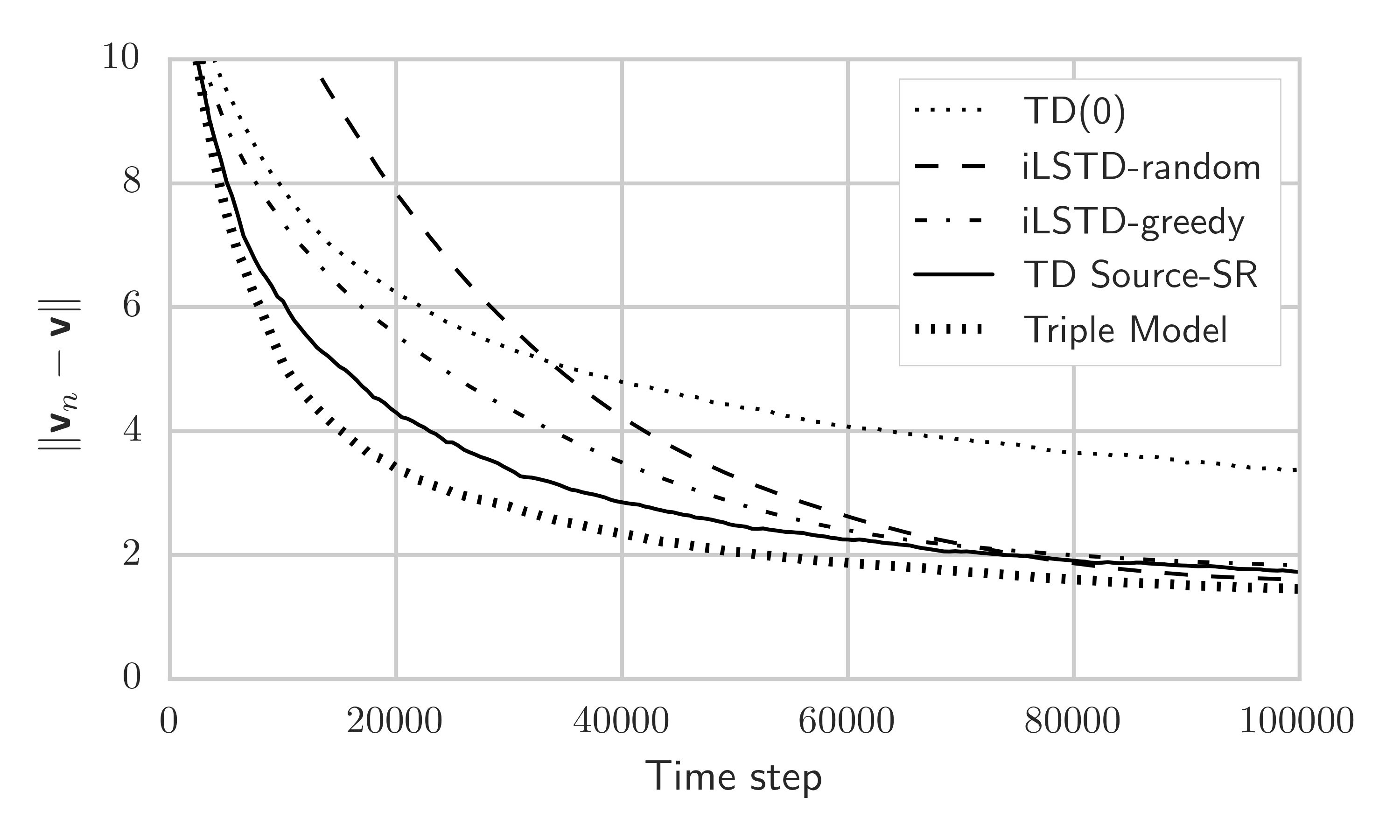}
\caption{Performance in 3D Gridworld using best tested annealing schedule. Compare to figure \ref{fig_convergence} (right).}
\label{fig_learned_vs_ideal}
\end{figure}

The results are shown in figure \ref{fig_learned_vs_ideal}, which is directly comparable to figure \ref{fig_convergence} (right). All four model-based approaches ended up in approximately the same place. Surprisingly, they performed comparably to source learning with ideal source traces, which tells us that a precisely accurate source map is not so important. Triple model learning had the fastest learning curve by a respectable margin, and both it and TD Source-SR learned faster than the two iLSTD algorithms (but note that only the best \textit{final} results are shown). By reducing the variance of updates, triple model learning outperforms even ideal source traces (final error of 1.45 vs 1.61). 

\section{Useful Properties of Source Traces} \label{section_benefits}

Setting aside any potential gains in learning speed, there are at least three properties that make the source model useful. 

\subsection{Time Scale Invariance}

Given some continuous process that can be modeled with a discrete-time MRP, there is a choice of how granular time steps should be; we may choose to model a time step as one second or as one tenth of a second. Source traces are invariant to the granularity in the following sense: if the credit that state $s_i$ receives for state $s_j$ is $[\textbf{S}]_{ij}$, this value will not change if the steps between $s_i$ and $s_j$ are defined with larger or smaller intervals (assuming $\gamma$ is adjusted appropriately). As noted by \citeauthor{dayan1993improving} \citeyear{dayan1993improving}, the source map ``effectively factors out the entire temporal component of the task.'' 

Time scale invariance means that source traces capture something about the underlying process that is independent of how we structure the state and transition model for that process. In particular, an agent equipped with a source map $\textbf{S}$ can use it to answer the following questions about cause and effect: ``What is likely to cause X?'' and ``What are the likely effects of Y?'' The answers could be used, in turn, for justification; i.e., as a step toward answering questions like: ``Why did you do X?'' or ``Why did Y happen?''

We leave further development on this topic to future work. 

\subsection{Reward Invariance}

Source traces are independent of the reward model. This makes them useful for transfer learning, for learning in non-stationary environments, and for evaluating hypotheticals.

The latter point---evaluating hypotheticals---is central to general intelligence. As this application is only relevant in control settings, however, we defer it to future work. 

On the former points---transfer learning and learning in non-stationary environments---source traces provide a method of quickly adapting to changing reward values. This was investigated from the lens of SR in the work of \citeauthor{dayan1993improving} \citeyear{dayan1993improving}, as well as in the follow-up works of \citeauthor{barreto2016successor} \citeyear{barreto2016successor}, \citeauthor{kulkarni2016deep} \citeyear{kulkarni2016deep}, \citeauthor{zhang2016deep} \citeyear{zhang2016deep}, and \citeauthor{lehnert2017advantages} \citeyear{lehnert2017advantages}. The only novelty offered by the backward-view in this context is the ability to propagate the impact of a reward change at a single state to the entire value space in $O(\vert S \vert)$ time. This ability may be useful in cases involving communication; for example, if a teacher informs the agent that they are misjudging the value of state $s_j$, that information can be used to directly update $\textbf{v}_0$. 

\subsection{Trajectory Independence}

Like eligibility traces, source traces distribute TD errors more than one step, thereby increasing sample efficiency. But unlike eligibility traces, source traces are \textit{independent of the current trajectory}. This means they can be used in cases where isolated experiences are preferred to trajectories. 

In particular, there are at least two reasons why we would prefer to sample experiences instead of trajectories from an experience replay or Dyna model. First, random sampling can alleviate problems caused by correlated data \cite{mnih2013playing}. Second, methods for prioritizing samples improve the efficiency by sampling experiences out of order \cite{schaul2015prioritized,moore1993prioritized}. In these cases, source traces may be used where eligibility traces are inapplicable. 

To demonstrate this benefit, we ran each of TD(0), TD(0) with ER, TD Source-SR, and TD Source-SR with ER at various \textit{fixed} learning rates in the 3D Gridworld environment. The replay memory had infinite capacity, and was invoked on every step to replay 3 past steps. For three ``target'' levels of error, we computed the number of real steps that it took for the average error (over the set of 30 3D Gridworlds) to fall below the target (in each case, at the best tested $\alpha$). The results in Table 1 show that switching from one-step backups to source backups during ER can speed up learning.

\begin{table}\label{table_exp_replay}
\centering
 \begin{tabularx}{\linewidth}{@{}l *3{>{\centering\arraybackslash}X}@{}} 
 \toprule
 & \multicolumn{3}{c}{Target $\Vert \textbf{v}_0 - \textbf{v} \Vert$} \\
 \cmidrule{2-4}
 & 3.0 & 2.0 & 1.5 \\
 \midrule
 TD(0) & 154 & 465 & 979 \\ 
 TD(0) w/ ER & 58 & 161 & 319 \\
 TD Source-SR & 58 & 158 & 335 \\
 TD Source-SR w/ ER & 41 & 100 & 189 \\
 \bottomrule
\end{tabularx}
\caption{Thousands of steps to reach target error values in 3D Gridworld when using experience replay and source traces}
\end{table}

\section{Conclusion} \label{section_conclusion}

This paper has developed the theory of source traces in the context of valuing tabular finite-state MRPs. Our contributions include the proposed source learning algorithm, a theorem for its convergence, the concept of partial source traces (and SR), the TD-Source and TD Source-SR algorithms, a comparison of direct methods and decomposition, the triple model learning algorithm, and a demonstration of the effectiveness of combining ER and source traces. While this work serves as a necessary foundational step, in most practical scenarios, we will be concerned with continuous, uncountable-state Markov decision processes (MDPs). In anticipation of future work, we conclude with a brief comment on the challenges posed by these settings.

For purposes of control, we note that $\textbf{S}$ changes as policy changes. To capture the same benefits in a control setting, one must either construct a higher order source function, learn a set of source maps (one for each policy), or rely on the error bound provided by the convergence theorem and the (not necessarily true) assumption that minor changes in policy entail only minor changes in the source map.

For purposes of approximation, note that extending source traces to propagate TD errors back to features entails the problem of agglomeration: since features, unlike states, are not isolated from each other (multiple features appear in the same state), collecting their historical accumulations into a single trace will erode the usefulness of its interpretability. It may therefore be more appropriate to construct a generative source model that captures a time-invariant distribution of past states. Such a model would maintain both its interpretability and usefulness for model-based learning. 

\footnotesize
\bibliography{st.bib}

\begin{thebibliography}{}

\bibitem[\protect\citeauthoryear{Barreto \bgroup et al\mbox.\egroup
  }{2016}]{barreto2016successor}
Barreto, A.; Munos, R.; Schaul, T.; and Silver, D.
\newblock 2016.
\newblock Successor features for transfer in reinforcement learning.
\newblock {\em arXiv preprint arXiv:1606.05312}.

\bibitem[\protect\citeauthoryear{Bradtke and Barto}{1996}]{bradtke1996linear}
Bradtke, S.~J., and Barto, A.~G.
\newblock 1996.
\newblock Linear least-squares algorithms for temporal difference learning.
\newblock In {\em Recent Advances in Reinforcement Learning}. Springer.
\newblock  33--57.

\bibitem[\protect\citeauthoryear{Dayan}{1993}]{dayan1993improving}
Dayan, P.
\newblock 1993.
\newblock Improving generalization for temporal difference learning: The
  successor representation.
\newblock {\em Neural Computation} 5(4):613--624.

\bibitem[\protect\citeauthoryear{Gehring}{2015}]{gehring2015approximatesr}
Gehring, C.
\newblock 2015.
\newblock Approximate linear successor representation (extended abstract).
\newblock {\em The 2nd Multidisciplinary Conference on Reinforcement Learning
  and Decision Making}.

\bibitem[\protect\citeauthoryear{Geramifard \bgroup et al\mbox.\egroup
  }{2007}]{geramifard2007ilstd}
Geramifard, A.; Bowling, M.; Zinkevich, M.; and Sutton, R.~S.
\newblock 2007.
\newblock i{LSTD}: Eligibility traces and convergence analysis.
\newblock In {\em Advances in Neural Information Processing Systems},
  441--448.

\bibitem[\protect\citeauthoryear{Geramifard, Bowling, and
  Sutton}{2006}]{geramifard2006incremental}
Geramifard, A.; Bowling, M.; and Sutton, R.~S.
\newblock 2006.
\newblock Incremental least-squares temporal difference learning.
\newblock In {\em Proceedings of the 21st national conference on Artificial
  intelligence-Volume 1},  356--361.
\newblock AAAI Press.

\bibitem[\protect\citeauthoryear{Jaakkola, Jordan, and
  Singh}{1994}]{jaakkola1994convergence}
Jaakkola, T.; Jordan, M.~I.; and Singh, S.~P.
\newblock 1994.
\newblock Convergence of stochastic iterative dynamic programming algorithms.
\newblock In {\em Advances in neural information processing systems},
  703--710.

\bibitem[\protect\citeauthoryear{Kemeny and Snell}{1976}]{kemeny1960finite}
Kemeny, J.~G., and Snell, J.~L.
\newblock 1976.
\newblock {\em Finite markov chains}.
\newblock Springer-Verlag, second edition.

\bibitem[\protect\citeauthoryear{Kulkarni \bgroup et al\mbox.\egroup
  }{2016}]{kulkarni2016deep}
Kulkarni, T.~D.; Saeedi, A.; Gautam, S.; and Gershman, S.~J.
\newblock 2016.
\newblock Deep successor reinforcement learning.
\newblock {\em arXiv preprint arXiv:1606.02396}.

\bibitem[\protect\citeauthoryear{Lehnert, Tellex, and
  Littman}{2017}]{lehnert2017advantages}
Lehnert, L.; Tellex, S.; and Littman, M.~L.
\newblock 2017.
\newblock Advantages and limitations of using successor features for transfer
  in reinforcement learning.
\newblock {\em arXiv preprint arXiv:1708.00102}.

\bibitem[\protect\citeauthoryear{Lin}{1992}]{lin1992self}
Lin, L.-H.
\newblock 1992.
\newblock Self-improving reactive agents based on reinforcement learning,
  planning and teaching.
\newblock {\em Machine learning} 8(3/4):69--97.

\bibitem[\protect\citeauthoryear{Mnih \bgroup et al\mbox.\egroup
  }{2015}]{mnih2013playing}
Mnih, V.; Kavukcuoglu, K.; Silver, D.; Rusu, A.~A.; Veness, J.; Bellemare,
  M.~G.; Graves, A.; Riedmiller, M.; Fidjeland, A.~K.; Ostrovski, G.; et~al.
\newblock 2015.
\newblock Human-level control through deep reinforcement learning.
\newblock {\em Nature} 518(7540):529--533.

\bibitem[\protect\citeauthoryear{Moore and
  Atkeson}{1993}]{moore1993prioritized}
Moore, A.~W., and Atkeson, C.~G.
\newblock 1993.
\newblock Prioritized sweeping: Reinforcement learning with less data and less
  time.
\newblock {\em Machine learning} 13(1):103--130.

\bibitem[\protect\citeauthoryear{Schaul \bgroup et al\mbox.\egroup
  }{2015}]{schaul2015prioritized}
Schaul, T.; Quan, J.; Antonoglou, I.; and Silver, D.
\newblock 2015.
\newblock Prioritized experience replay.
\newblock {\em arXiv preprint arXiv:1511.05952}.

\bibitem[\protect\citeauthoryear{Sutton and
  Barto}{1998}]{sutton1998reinforcement}
Sutton, R.~S., and Barto, A.~G.
\newblock 1998.
\newblock {\em Reinforcement learning: An Introduction}.
\newblock The MIT Press, Cambridge.

\bibitem[\protect\citeauthoryear{Sutton}{1988}]{sutton1988learning}
Sutton, R.~S.
\newblock 1988.
\newblock Learning to predict by the methods of temporal differences.
\newblock {\em Machine learning} 3(1):9--44.

\bibitem[\protect\citeauthoryear{Sutton}{1990}]{sutton1990integrated}
Sutton, R.~S.
\newblock 1990.
\newblock Integrated architectures for learning, planning, and reacting based
  on approximating dynamic programming.
\newblock In {\em Proceedings of the seventh international conference on
  machine learning},  216--224.

\bibitem[\protect\citeauthoryear{Tsitsiklis and
  Van~Roy}{1997}]{tsitsiklis1997analysis}
Tsitsiklis, J.~N., and Van~Roy, B.
\newblock 1997.
\newblock Analysis of temporal-diffference learning with function
  approximation.
\newblock In {\em Advances in neural information processing systems},
  1075--1081.

\bibitem[\protect\citeauthoryear{White}{1996}]{white1996temporal}
White, L.~M.
\newblock 1996.
\newblock {\em Temporal difference learning: eligibility traces and the
  successor representation for actions.}
\newblock Univ. of Toronto.

\bibitem[\protect\citeauthoryear{Zhang \bgroup et al\mbox.\egroup
  }{2016}]{zhang2016deep}
Zhang, J.; Springenberg, J.~T.; Boedecker, J.; and Burgard, W.
\newblock 2016.
\newblock Deep reinforcement learning with successor features for navigation
  across similar environments.
\newblock {\em arXiv preprint arXiv:1612.05533}.

\end{thebibliography}
\bibliographystyle{aaai}
\end{document}